\pdfoutput=1
\documentclass[11pt]{article}

\usepackage[preprint]{acl} 
\usepackage{times} 
\usepackage{latexsym}
\usepackage[T1]{fontenc} 
\usepackage[utf8]{inputenc} 
\usepackage{microtype} 
\usepackage{inconsolata} 

\usepackage{amsmath}
\usepackage{graphicx}
\usepackage{color}
\usepackage{booktabs}
\usepackage{multirow}
\usepackage{enumitem}

\title{Why do LLaVA Vision-Language Models Reply to Images in English?}

\author{
  \textbf{Musashi Hinck\thanks{First author.}\textsuperscript{1}},
 \\ 
  \textbf{Carolin Holtermann\thanks{Equal contribution, ordered alphabetically.}\textsuperscript{2}}
  \textbf{Matthew Lyle Olson\footnotemark[2]\textsuperscript{1}},
  \textbf{Florian Schneider\footnotemark[2]\textsuperscript{2}}
\\
  \textbf{Sungduk Yu\textsuperscript{1}},
  \textbf{Anahita Bhiwandiwalla\textsuperscript{1}},
\\
  \textbf{Anne Lauscher\textsuperscript{2}}
  \textbf{Shaoyen Tseng\textsuperscript{1}},
  \textbf{Vasudev Lal\textsuperscript{1}},
\\
  \textsuperscript{1}Intel Labs,
  \textsuperscript{2}University of Hamburg,
\\
  \small{
    \textbf{Correspondence:} \href{mailto:musashi.hinck@intel.com}{musashi.hinck@intel.com}
  }
}

\begin{document}
\maketitle
\begin{abstract}

We uncover a surprising multilingual bias occurring in a popular class of multimodal vision-language models (VLMs). Including an image in the query to a LLaVA-style VLM significantly increases the likelihood of the model returning an English response, regardless of the language of the query.
This paper investigates the causes of this loss with a two-pronged approach that combines extensive ablation of the design space with a mechanistic analysis of the models' internal representations of image and text inputs.
Both approaches indicate that the issue stems in the language modelling component of the LLaVA model. Statistically, we find that switching the language backbone for a bilingual language model has the strongest effect on reducing this error. Mechanistically, we provide compelling evidence that visual inputs are not mapped to a similar space as text ones, and that intervening on intermediary attention layers can reduce this bias.
Our findings provide important insights to researchers and engineers seeking to understand the crossover between multimodal and multilingual spaces, and contribute to the goal of developing capable and inclusive VLMs for non-English contexts.


\end{abstract}

\section{Introduction}
\label{intro}
Language fidelity in large language models (LLMs) refers to whether the model replies in the same language as it was queried in. 
While seemingly a simple task for humans, models with multilingual capabilities will often bias towards English replies especially for queries in low-resource languages \cite{holtermann2024evaluating}.
In this work, we identify a surprising but parallel pathology in LLaVA-style VLMs \cite{liu2023llava}: when prompted with a multimodal query that includes an image, the model is more likely to reply in an incorrect language with respect to the query language. We term this Image-induced Fidelity Loss (IFL).

In the first part of the paper, we define this phenomenon and demonstrate the \textit{extent} of this problem in existing LLaVA-style models. Using a collection of 7740 evaluation tasks drawn from three VQA benchmarks and spanning fourteen languages, we experimentally show that adding an image to the query to a LLaVA model causes the probability of the response being in the correct language to decrease between $6\%$ and $53\%$.

To gain a holistic understanding of IFL, we employ a two-pronged approach that combines design space exploration with introspective analysis of the model's internal representations. First, we conduct a macro-level analysis by systematically ablating the design space of LLaVA-style models and train multiple variations. This allows us to statistically measure the impact of each design choice on the model's propensity for generating linguistically misaligned responses. Second, we complement this macro-level perspective with a micro-level investigation of the model's intermediary representations. By studying the clustering patterns of vision tokens in relation to tokens from different languages, and directly intervening on hidden states within the language transformer layers, we gain insights into the internal dynamics that give rise to the observed phenomenon.

The synthesis of these two levels of analysis – exploring the design space at a statistical level and probing the model's internal representations – provides a comprehensive characterization of the factors influencing cross-lingual response generation in multimodal models. This multi-level approach enables us to identify the key design choices and representational properties that contribute to this phenomenon.

Therefore, our contributions are as follows: 
\begin{itemize}
    \item We systematically demonstrate IFL on multiple tasks over a wide range of languages for modern VLMs.
    \item We conduct a large scale analysis of the design choices in training VLMs, enabling robust statistical results for interpreting training architecture decisions.
    \item We perform detailed representation analysis to find the IFL problem is localized within the language model.
\end{itemize}

\section{Related Work} 
\label{sec_related}

\subsection{Global Inclusion via Multilingual Models} Aiming towards globally inclusive language technologies, much of natural language processing research has focused on multilingual and cross-lingual models~\cite{conneau-etal-2018-xnli}. Starting from static embedding models~\citep[cf.,][]{ruder_survey_2019} and smaller multilingual pretrained transformer models like mBERT~\citep{devlin-etal-2019-bert} and XLM-R~\citep{conneau-etal-2020-unsupervised}, much of multilingual NLP has shifted to multilinguality in instruction-tuned LLMs. Here, researchers either focus on training multilingual LLMs~\citep{lai-etal-2023-okapi}, or, given that explicitly open multilingual chat models are still rare, aim towards understanding multilinguality of models intended for English use only~\citep{blevins-zettlemoyer-2022-language}. 
Moreover, the vast adoption of instruction-tuned models has transitioned their focus from task-specific responses to generating more natural language outputs. This allows models to answer more flexibly, unbound by rigid response frameworks, and autonomously choose the language of their responses. Despite this advancement, existing benchmarks primarily evaluate the accuracy of answers without adequately assessing the models' fidelity to the language used \cite{holtermann2024evaluating}. However, given the multilingual use of these models \cite{zhao2024wildchat}, it is crucial to ensure they can respond accurately in the appropriate language to foster inclusivity.

\subsection{Efficient Integration for Multimodal Understanding}
Humans interact with the world through multiple channels. Accordingly, many AI researchers explored how to integrate multiple modalities, particularly vision and language, into a single model~\citep[e.g.,][]{kim_vilt_2021,wang_simvlm_2022}. As an alternative to efforts that focused on models specific to particular tasks~\citep{brooks_instructpix2pix_2023}, general-purpose vision-language models emerged. Given that pretraining larger models~\citep[as in][\emph{inter alia}]{kim_vilt_2021,radford_learning_2021} became prohibitively expensive, researchers moved to employing readily available encoders, keeping those (partially) frozen, and mapping them through learned projections~\cite{li_blip-2_2023, merullo_linearly_2023}. For instance, \citet{manas_mapl_2023} employ a transformer mapping network, and \citet{eichenberg-etal-2022-magma} rely on adapters. In this work, we focus on  efficient and thus, accessible methods for the integration of instruction-tuned LLMs, with LLaVA as a popular representative~\citep[][]{liu2023llava}.

\subsection{Multimodal Understanding and Multilinguality}
The resource efficiency of LLaVA has already enabled several works creating multilingual LLaVA models. \citet{andersland2024amharic} trains an Amharic LLaVA model by employing methods previously used for languages with data scarcity such as machine translation, dataset augmentation and dataset expansion.
Similarly \citet{shin2024x} train an English-Korean-Chinese trilingual LLaVA model using vocabulary expansion and a trilingual pretraining data mixture.
Looking at the failures of VLMs, \citet{song2024missing} focus on three key aspects: multilinguality, complex reasoning and multimodality. To address these issues, they propose three interventions that show significant improvements in model performance. The translate-test approach enhances multilingual processing, visual programming simplifies complex reasoning, and image captioning boosts multimodal understanding.
In our approach, we specifically focus on developing mechanisms to understand IFL by analyzing the linguistic and visual encoder integration, aiming to explain language fidelity issues even in the face of complex multimodal inputs.




\section{Image-induced Fidelity Loss}
\label{sec_ifl}

The phenomenon of interest in this study is the change in fidelity seen when adding an image to the input to a VLM. We refer to this as \textit{image-induced fidelity loss} (IFL). We argue that this phenomenon is surprising, given that visual inputs should be language-agnostic (barring the cultural-linguistic context associated with the image) and therefore orthogonal to the language of the response. 

\subsection{Experimental Design}
Following this intuition of orthogonality, we design an experiment comparing the response language of VLMs when prompted with two ``semantically equivalent'' inputs: an image-plus-text query, versus the same query with the image replaced with a text description of its contents. We collect these image-text pairs from the three multilingual VQA benchmarks: MaXM \citep{changpinyo2023maxm}, PALO-LLaVAW \citep[hereafter LLaVAW]{maaz2024palo} and ViSIT \citep{bitton2023visit}. These are summarized in Table \ref{tab:dataset_summary}.

The text descriptions of the images are drawn from the respective datasets, and in the case of ViSIT are generated conditional on the task instruction and verified by human annotators. In total, for each model we collect $15480$ responses spanning fourteen languages ($7740$ per treatment condition; Table \ref{tab:lang_count}).

\begin{table}[t!]
\centering
    \begin{tabular}{llrr}
        \toprule
        \textbf{Dataset} & \textbf{MM} & \multicolumn{1}{l}{\textbf{\#Langs.}} & \multicolumn{1}{l}{\textbf{Size}}  \\
        \midrule
        PALO-LLaVAW     & yes & $10$ & $600$     \\
        MaXM            & yes & $7$ & $2142$     \\
        ViSIT           & yes & $10$ & $5740$     \\
        MultiQ          & no & $119$ & $27400$    \\
        \bottomrule
    \end{tabular}
\caption{Overview of the employed datasets, indicating multimodality (MM), number of languages (\#Langs.), and total number of observations (including parallel tasks repeated between languages).}
\label{tab:dataset_summary}
\end{table}

\begin{table}[tphb]
\centering
\begin{tabular}{lr|lr}
\toprule
\textbf{Language} & $N$ & \textbf{Language} & $N$ \\
\midrule
Chinese (zh)  & 862 & Japanese (ja) & 585 \\
Hindi (hi)    & 845 & Spanish (es)   & 585 \\
English (en)  & 842 & German (de)   & 525 \\
Hebrew (he)   & 805 & French (fr)   & 324 \\
Thai (th)     & 793 & Romanian (ro) & 284 \\
Arabic (ar)   & 585 & Russian (ru)  &  60 \\
Bengali (bn)  & 585 & Urdu (ur)     &  60 \\
\bottomrule
\end{tabular}
\caption{Number of tasks per language in the datasets we study. ISO 639 language codes are provided in parentheses.}
\label{tab:lang_count}
\end{table}

\subsection{Measurement Strategy}
We use the GlotLID model \citep{kargaran-etal-2023-glotlid} to predict the language of the model output. If the predicted language of the response is the same as the query\footnote{See supplementary materials for notes on manual postprocessing of GlotLID outputs.}, we score the response as having a predicted fidelity of $1$. If the languages do not match, then we score the response as having a predicted fidelity of $0$.
During the process, we observed non-random errors in the GlotLID predictions, such as having a lower accuracy on shorter texts. We correct for this bias in our downstream statistical analyses using the design-based supervised learning framework \citep[DSL]{egami2023dsl}. DSL leverages a small number of randomly sampled expert annotations to correct for bias in downstream estimators caused by imperfect proxy measures. We manually label a stratified random sample of $1000$ examples to use as our gold standard. The debiased results can be interpreted as being the results that would have been obtained if we had used expert annotation for all datasets. We provide details of the sampling weights and annotation method in the supplementary materials.

\subsection{Prevalence of IFL}
To assess the prevalence of the IFL problem in existing LLaVA models, we apply the above experiment to four popular LLaVA-style VLMs: LLaVA-v1.5-7b, LLaVA-v1.5-13b \citep{liu2023llava}, BakLLaVA \citep{skunkworksai2023bakllava1} and LLaVA-Gemma-2B \citep{hinck2024llavagemma}.\footnote{We limit our scope to LLaVA-v1.5 models because the exact data mixture and training architecture for the newer v1.6 models has not been made public at the time of writing. Given our stated aim to explore the effects of the training decisions, we cannot provide insights into the newer models.}

\begin{figure}
    \centering
    \includegraphics[width=\linewidth]{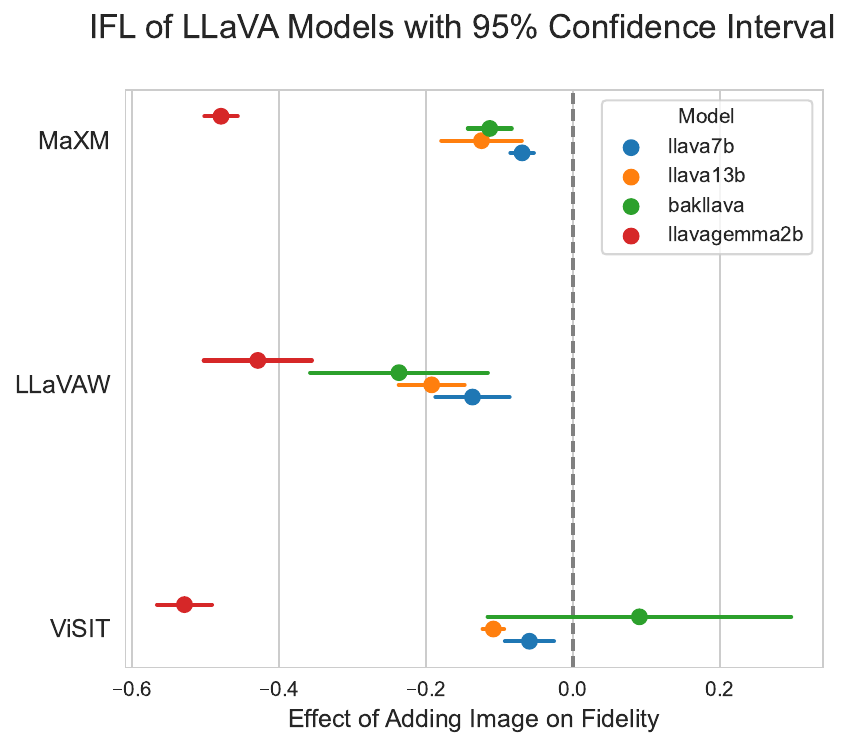}
    \caption{Effect of adding image to query on response fidelity with 95\% confidence intervals. All estimates are aggregated over languages within the benchmark and debiased using the DSL estimator.}
    \label{fig:ifl_orig}
\end{figure}

Figure \ref{fig:ifl_orig} shows the estimated effect and 95\% confidence interval of adding an image on fidelity for each model and benchmark. Several patterns emerge: for LLaVA-v1.5-7b, adding an image to the query causes fidelity to fall between 5.9 percentage points (pp) on ViSIT to 13.7 pp on LLaVAW. The effect on LLaVA-v1.5-13b is even greater, ranging from a 10.8 pp (ViSIT) to 19.2 pp (LLaVAW) decrease in fidelity. LLaVA-Gemma-2b displays the IFL, ranging from a 42.9 pp decrease on LLaVAW to a 52.9 pp decrease on ViSIT. With the exception of BakLLaVA on ViSIT, all effects are statistically significant at an alpha of $0.95$.

Collectively, these results provide concrete evidence of a systematic issue: LLaVA models are more likely to reply in the wrong language when the user includes an image in the query. The remainder of this paper explores the source of this issue.




\section{Effects of Design Choices}
\label{sec_design}

The LLaVA architecture combines a pretrained vision encoder and language model by using a small multi-layer perceptron (MLP) to project the penultimate hidden states of the vision encoder into the input embedding space of the language model \citep{liu2023llava}. This architecture is then fine-tuned with two stages of training. In the first, the vision and language models are frozen and the projection MLP is trained on 558k image-caption pairs. In the second, the vision encoder is kept frozen and the projection MLP and language model are trained on 665k visual instruction-following and examples \citep{liu2023llava}.

\subsection{Design Space}
In constrast to related works that study the effect of architectural decisions in the VLM design space \citep{karamcheti2024prismatic}, we focus our analysis on the effect of the choice of pretrained models and training data by holding the architecture constant.

LLaVA uses Vicuna-v1.5-7b \citep{zheng2023judging} as the language model, CLIP \citep{radford_learning_2021} as the vision encoder and English for more than 99\% its training examples. There are reasons \textit{a priori} to think that any of these decisions could induce an ``English bias'' in the model. Vicuna is published as an English-language LLM trained primarily on English-language examples. The captions used to train the CLIP vision encoder are filtered for non-English texts \citep[p.3]{radford_learning_2021}, meaning that the representations produced by CLIP may be ``biased'' towards English language representations of visual data. Finetuning the model with primarily English data may ``teach'' the language model to reply to visual inputs from the vision encoder/MLP in English.

We ablate each of these design choices individually while holding architectural features constant to disentangle their effects. For our experiments, we focus on Chinese and German because these are languages for which there is an LLM at a similar size and architecture to Vicuna-7b that is not directly finetuned from Vicuna-7b. For Chinese, we use the Yi-6b-chat, a 6B-parameter LLM trained from scratch on a bilingual Chinese-English data mixture \citep{ai2024yi}. For German, we use LeoLM-7b-chat, a 7B-parameter LLM finetuned from Llama-2 \citep{touvron2023llama} on 65B German tokens \citep{pluster2023leolm}. For the vision encoder, we test the effect of substituting CLIP for DINOv2 \citep{oquab2024dinov2} because the latter is trained using a self-supervised training objective that does not incorporate language, while holding the architecture constant. We use NLLB-1.7-distilled \citep{nllbteam2022language} to machine translate all $\sim$1.2M training observations used in the LLaVA training data into Chinese and German. We discuss the implications of the errors in the machine translation pipeline in a subsequent section.

\begin{table}
\centering
\begin{tabular}{ll}
\toprule
\textbf{Axis} & \textbf{Options} \\ \midrule
LLM & 
\begin{tabular}[t]{@{}l@{}}
Vicuna-v1.5-7b \citep{zheng2023judging}\\ 
Yi-6b-chat \citep{ai2024yi} \\ 
Leo-7b-chat \citep{pluster2023leolm}
\end{tabular} \\ \hline
Vision & 
\begin{tabular}[t]{@{}l@{}}
CLIP \citep{radford_learning_2021}\\ 
DINOv2 \citep{oquab2024dinov2}
\end{tabular} \\ \hline
Data & 
\begin{tabular}[t]{@{}l@{}}
English \citep{liu2023llava} \\ 
Chinese \\ 
German
\end{tabular} \\
\bottomrule
\end{tabular}
\caption{Summary of design space. Note that Yi is only combined with English and Chinese, and Leo with English and German.}
\label{tab:design_summary}
\end{table}

This design yields a total of fourteen combinations (Table \ref{tab:design_summary}), which we trained on using 8 $\times$ A6000 Nvidia GPU nodes on an internal cluster. All designs used the same training parameters as the original LLaVA-v1.5-7B model. We provide further training details in the supplementary materials.

\subsection{Design Effects}
For each set of experiments (Yi/Chinese and Leo/German), we measure the effect of training choices on IFL using the following regression model with first-order interactions:

\begin{align*}
\text{Fidelity} = \beta_0 &+ \beta_1\text{Image} \\
                  &+ \beta_2\text{Image} \times \text{Lang. Model} \\
                  &+ \beta_3\text{Image} \times \text{Vision Model} \\
                  &+ \beta_4 \text{Image}\times \text{Data Lang.} + \epsilon
\end{align*}

\noindent where:

\begin{itemize}[itemsep=2pt, parsep=2pt, topsep=0pt, partopsep=0pt]
    \item `Fidelity' is a binary indicator for whether a completion to a query in the target language (Chinese or German) is in the correct language
    \item $\beta_0$ is a constant intercept term that captures the average fidelity of the reference class (LLaVA-v1.5-7b)
    \item $\beta_1 \text{Image}$ captures the effect of adding an image to the query on fidelity (i.e., IFL) for the reference class
    \item $\beta_2\text{Image}\times\text{Lang. Model}$ is an interaction term capturing the change in IFL associated with changing the LLM backbone from Vicuna to Yi or Leo
    \item $\beta_3\text{Image}\times\text{Vision Model}$ captures the change in IFL associated with changing the vision backbone from CLIP to DINOv2
    \item $\beta_4\text{Image}\times\text{Data Lang.}$ captures the change in IFL associated with changing the training data from English to the target language.
    \item $\epsilon$ is an error term
\end{itemize}

Coefficients $\beta_2$, $\beta_3$ and $\beta_4$ with the corresponding 95\% confidence interval are reported in the left-hand side of Table \ref{tab:design_effects}. We see similar patterns for both languages. Changing the language model from Vicuna to Yi/Leo improved the performance of the model, reducing IFL by 17 and 7 pp for Chinese and German respectively. Changing the vision encoder from CLIP to DinoV2 worsened IFL, increasing it by 20 and 11 pp respectively. Changing the training data language worsened IFL considerably, increasing it by 16 and 37 pp respectively.


\begin{table}[t]
\centering
\setlength{\tabcolsep}{4pt} 
\begin{tabular}{c|rr}
\toprule
 \textbf{Model} & \multicolumn{1}{c}{\textbf{IFL}} & \multicolumn{1}{c}{\textbf{Accuracy}} \\
\midrule
&\multicolumn{2}{c}{\underline{\hspace{6.1em}Chinese\hspace{6.1em}}}\\
LLM & $0.17$ [0.15, 0.19] & 0.21 [-0.07, 0.50] \\
VE & -0.20 [-0.22, -0.18] & 0.15 [-0.13, 0.43] \\
Data & -0.16 [-0.17, -0.14] & 0.01 [-0.27, 0.30] \\
\midrule
&\multicolumn{2}{c}{\underline{\hspace{6.1em}German\hspace{6.1em}}}\\
LLM & 0.07 [0.04, 0.10] & 0.28 [-0.35, 0.91] \\
VE & -0.11 [-0.15, -0.08] & -0.10 [-0.73, 0.53] \\
Data & -0.37 [-0.40, -0.33] & -0.24 [-0.87, 0.40] \\
\bottomrule
\end{tabular}
\caption{Average effect of design choice on IFL (left-hand column) and Accuracy (right-hand column for Chinese (top) and German (bottom). Each value is reported with the 95\% confidence interval in brackets. Row LLM indicates the effect of changing Vicuna to a LLM that is capable in the target language. Row VE indicates the effect of changing CLIP to DINOv2. Row Data indicates the effect of machine translating the training data.}
\label{tab:design_effects}
\end{table}

\subsection{Effect on Accuracy}
Although the above section provides insights into reducing IFL, how do these design decisions affect the factual accuracy of responses? To measure this, we used GPT-4o \citep{openai_gpt4o_2024} to generate zero-shot predictions of the accuracy. Our GPT-4o prompt gave the question, dataset ground truth (where available) and model completion and asked if the completion is correct given the question and ground truth label. We then used the DSL procedure to debias these evaluations, whereby the authors manually annotated $1000$ observations to provide a gold standard.

The right-hand column of table \ref{tab:design_effects} displays the effect of each design decision on the accuracy of responses in the target language. The results are uniformly inconclusive, with all coefficients not statistically distinct from $0$. We interpret these results to mean that the effects of design choices on accuracy exhibit high variance. This implies that a larger number of expert annotations in these subsets is necessary to achieve sufficient statistical power to detect these small or noisy effects.



\section{Locating the Cause of IFL} 
\label{sec_interpret}
\begin{figure*}[ht]
\centering
\includegraphics[width=0.32\textwidth]{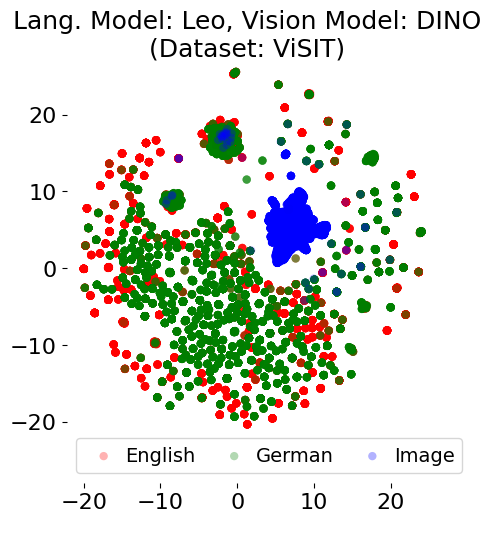}
\includegraphics[width=0.32\textwidth]{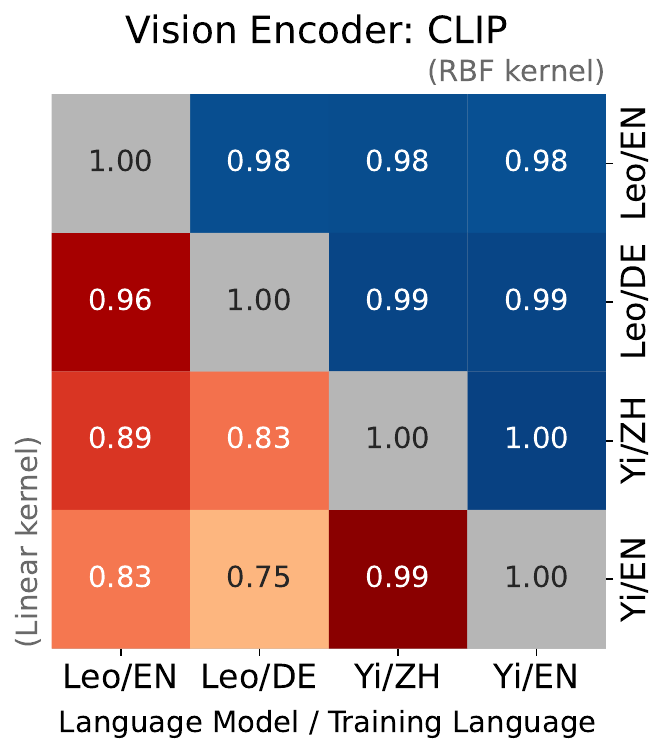}
\includegraphics[width=0.32\textwidth]{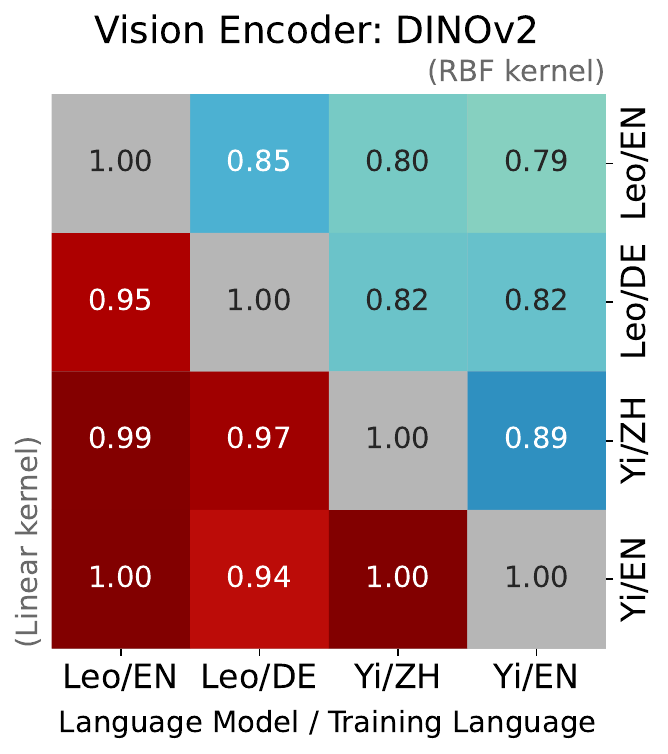}
\caption{ \textbf{(LEFT):} UMAP visualization of image and text embeddings from a multimodal language model. Image embeddings are shown clustering distinctly from text embeddings, indicating a unique separation in the latent space. This segregation highlights potential areas of focus for addressing fidelity loss in multimodal communication.
\textbf{(Center and Right):}Centered Kernel Alignment (CKA) heatmap showing the similarity of vision embeddings across two differently trained language models. CKA based on a linear kernel is shown in the lower triangle; CKA based on an RBF kernel is shown in the upper triangle. The heatmap reveals a high degree of similarity in how vision data is embedded, regardless of the language model’s architecture or training data specifics. This uniformity suggests that the method of integrating visual data into language models is a critical factor affecting fidelity.}
\label{fig:umap_cka}
\end{figure*}

The previous experiments provide macro-level evidence that the LLM influences IFL based on the inputs and outputs of the models, but does not shed any light on what happens inside the model.
In this section, we explore whether IFL is primarily attributable to any one component-- finding the problem to reside in the LLM. And we find that with a strikingly simple training-free mechanistic intervention in the LLM, we can substantially reduce IFL.

\subsection{UMAP Analysis of Embedding Spaces}

To gain a qualitative understanding for how image embeddings interact with text embeddings in the input space, we employ Uniform Manifold Approximation and Projection (UMAP) for dimensionality reduction \cite{mcinnes2018umap}. UMAP is a non-linear dimensionality reduction technique that preserves global data structure, making it ideal for visualizing high-dimensional data.

In our experiment, embeddings from text inputs and image tokens are jointly visualized using UMAP. Figure \ref{fig:umap_cka} (left) illustrates that image embeddings cluster distinctly from text, demonstrating a demarcated separation in the latent space. This segregation suggests image embeddings predominantly occupy a unique region of the embedding space, indicating they are not directly embedding in the same area as any particular language.

\subsection{CKA Analysis of Vision Embeddings}
To further understand what is happening with the image embeddings, we use Centered Kernel Alignment (CKA) to measure the similarity of internal representations across differently trained models \cite{kornblith2019similarity}. CKA measures the similarity between two sets of data by comparing kernel matrices, which transform data into a high-dimensional space. A CKA score close to 1 indicates high similarity between datasets, while a score near 0 suggests low similarity. We use CKA in order to measure how the vision embeddings compare between two seperately trained VLMs: LLaVA-Yi trained in Chinese and LLaVA-Leo trained in German.

Figure \ref{fig:umap_cka} (center and right) shows a surprising result that vision embeddings maintain a consistent structure in the latent space across various models, regardless of the language backbone or the training data specifics. This consistency supports the finding in the UMAP visualization that image embeddings are in their own region of the input space. Additionally, it is the language model's responsibility to interpret these out-of-distribution embeddings in the language it determines best; it is not the case that the MLP adaptor model is placing the image embeddings near a particular language. Next, we explore how to encourage the LLM to remedy the IFL problem.

\subsection{Mechanistic Intervention: Experiment}
An exciting nascent field in LLM research is Mechanistic interpretability, in which the ability to explicitly steer LLMs has recently been shown. This field shows great promise in modifying language model outputs, as demonstrated in recent studies that ablate refusals from tuned LLMs \cite{arditi2024refusal} and the isolation of interpretable attributes \cite{openAI2024scalingSAE, templeton2024scaling}.
 
Given this ability to manually modify an LLM without training nor modifying the weight, we propose an intervention approach to reduce IFL by directly modifying the model's intermediate representations at runtime. We use a strikingly simple steering mechanism, using just 1 text example per language, and achieve significant reduction in IFL. 

Our steering mechanism works by computing a language attribute $a_{lang}$ in an intermediate layer, then applying that attribute to every generated token. The attribute is computed as follows:
$$
a_{lang} =  LLM_l(x_{lang}) - LLM_l(x_{en})
$$

where $LLM_l$ represents the output at layer $l$, $x_{en}$ is the sentence ``Describe this image in detail.'', and $x_{lang}$ is the translated version of that sentence.

During inference, this direction is added to the output of layer $l$, effectively steering the model’s behavior towards the desired language:
$$
LLM*_l = LLM_l(o_{l-1}) + a_{lang} 
$$
where $o_{l-1}$ is the output of the previous layer and $LLM*_l$ is the new, intervened layer. For our experiments layer $l$ is selected to be partway through computation at one third depth (e.g., layer 10 out of 30).

\subsection{Mechanistic Intervention: Results}
The application of this mechanistic intervention has shown significant improvements in the fidelity of the model’s responses across various languages. The quantitative improvements are displayed in Table \ref{tab:mechint_results}, which presents a comparative analysis of fidelity metrics before and after the intervention. We find large relative improvements in performance across pretrained VLMs. Furthermore, we observe an improvement of IFL over the base LLM without image input. 

\begin{table}[t]
\centering
\setlength{\tabcolsep}{4.9pt}
\begin{tabular}{lrrrr}
\toprule
             \textbf{Model} & \textbf{IFL}    & \begin{tabular}[r]{@{}r@{}}\textbf{IFL +} \\ \textbf{Remedy}\end{tabular} & \textbf{Diff.} & \begin{tabular}[c]{@{}l@{}}\textbf{Relative} \\ \textbf{Increase}\end{tabular} \\
\midrule
llava7b      & -0.085 & -0.030                                                   & 0.055 & 65\%                                                            \\
llava13b     & -0.175 & -0.103                                                  & 0.073 & 42\%                                                            \\
bakllava     & -0.073 & 0.098                                                   & 0.170  & 233\%                                                           \\
 \begin{tabular}[c]{@{}l@{}}llava- \\ gemma2b\end{tabular} & -0.681 & -0.513                                                  & 0.168 & 25\%                                                            \\ 
 \bottomrule
\end{tabular}
\caption{Fidelity improvements by using mechanistic intervention (Remedy). Across all pretrained models, we find significant reduction in IFL by interventing on the LLM's intermediate layer. A full breakdown is available in the supplement.}
\label{tab:mechint_results}
\end{table}

While this experiment does require knowledge about which language attribute to select, it does clearly confirm our hypothesis that the LLM is responsibility for IFL. Moreover, the successful application of a targeted mechanistic intervention highlights the potential of this approach to effectively address and resolve issues related to IFL, a future area of research we are interested in exploring further.

\section{Implications} 
\label{sec_implications}
\subsection{Multilingual Multimodal Understanding}
Our results provide strong evidence in favor of the hypothesis that LLaVA models do not treat visual inputs orthogonally with respect to multilinguality, and this bias occurs in the language backbone of the VLM.
In the first part of our analysis, we demonstrate the prevalence of IFL across different LLaVA models and languages. The second part of our analysis provides a controlled comparison of the effects of changing the linguality of the language backbone, the pretraining objective of the vision encoder, and the language of the training data.

We find that substituting the monolingual Vicuna language backbone for a bilingual Yi or Leo backbone reduces IFL in Chinese and German. In their exploration of the latent representations of multilingual inputs in Llama-2 models, \citet{wendler2024llamas} find that the abstract ``concept space'' of these LLMs lies closer to English than other languages. Correspondingly, if we think of LLaVA as instructing LLMs to map visual inputs to latent linguistic representations, an explanation for the positive effect of using a bilingual backbone may be that the higher proportion of the target language in the training data mix may reduce the extent to which visual concepts are mapped to spaces closer to English.

This interpretation is supported by our investigations of the intermediate representations of visual and text inputs to our suite of ablated LLaVA models. The UMAP and CKA results show that visual inputs are located in a separate space to textual inputs, and that this is remarkably consistent across training configurations. This indicates that LLaVA ``instructs'' the language backbone to interpret visual inputs, instead of learning to map visual inputs to a linguistic semantic space. Further evidence of the multimodal fusion occurring at an intermediary layer of the language model is shown in our mechanistic intervention analysis, where we show that the bias can be partially corrected through intervention on these layers.

We also find that machine translating the training data language and changing the vision encoder to a language-agnostic one do not mitigate IFL. These null results should not necessarily be interpreted as negative ones; alternative explanations include limitations in performance stemming from the quality of the machine translated data, or there being a specific incompatibility between DINOv2 and the language backbones used.

\subsection{Building Robust Multilingual Multimodal Models}
In terms of implications for the design of \textit{multilingual} multimodal models, our research suggests that focus on improving the multilinguality of LLMs provides a productive way forward for building more robust multilingual VLMs. This finding is largely in line with the recent work on multimodal models, where progress has been fuelled by the increased availability of permissively licensed and capable pretrained LLMs. The mechanistic intervention results also suggest that effective and low-cost corrections to particular forms of bias (such as IFL) are a fruitful research direction.

\section{Future Work}
\label{sec_future}
As the adoption of generative models proliferates, multilingual multimodal models that can service multiple regions will become an attractive solution.  However, further efforts can be made to create more effective multilingual VLMs.
One possible approach is the consideration of language fidelity (image-induced or not) during the training process through a mix of multilingual data and/or language-controlling instructions for in-context learning. 
Other in-depth approaches may include the design of better model architectures that integrate multitmodal representations at a suitable semantic level to prevent language bias. 
The phenomenon of IFL may also serve as a motivation to better understand the semantic role of visual representations within VLMs and how they are interpreted by the base language model.

We are particularly interested in understanding and improving the mechanistic intervention. We think this is a promising area of research, and that our simple approach could be improved either by automatic attribute selection conditioned on internal representation or better construction of the attribute. We also would like to study the broad class of fidelity loss in single modality models and to what extent mechanistic interventions are useful in those scenarios.

\section{Conclusion}
In this work, we systematically examined the phenomenon of Image-induced Fidelity Loss in multimodal language models. Our analyses reveal that the addition of visual content to queries can unexpectedly bias the language output of these models, often resulting in responses in an incorrect language. Through a combination of empirical evaluations and introspective techniques, we quantified the extent of IFL across various VLM architectures and explored how different design choices impact language fidelity. Our findings are further substantiated by micro analyses within the models, illustrating the intricate interactions between image and text embeddings. Our introduction of a targeted mechanistic intervention demonstrates a potential method to mitigate IFL, indicating IFL is localized within the LLM itself, and suggesting a pathway for future enhancements to VLMs. Our work contributes to the broader understanding of multimodal systems and offers actionable insights for developing more robust and linguistically accurate AI technologies.

\section{Limitations}
\label{sec_limits}

\subsection{Assumptions and Scope}
When we debias our results with DSL, we assume that the expert annotated gold standard is in correct. Nevertheless, it is possible that the authors made mistakes during the data annotations, especially when identifying languages in unfamiliar scripts. The implication of this is that the bias-corrected results should be interpreted as ``the answer we would get if we had manually annotated all of the examples'' instead of ``the truth''.

The analyses in Section \ref{sec_interpret} in Section \ref{sec_implications} contain a degree of speculation that we want to communicate clearly. The visible separation between image and text embeddings seen in the UMAP visualizations do not definitively prove that the inputs are mapped to distinct ``semantic'' spaces, and are not interpreted as such. Rather our takeaway is the lack of clustering between the visual tokens and a particular language that would be suggestive the projection layer learns to map images to a particular language.

The focus of this paper -- IFL -- could be framed as a form of model bias, and we sometimes describe it as such. We believe our definition of IFL is clearly articulated and operationalized: the decrease in linguistic fidelity in a VLM caused by adding an image to the input. We think that this constitutes a form of bias in the sense that it is a deviation from an intended/desirable output (i.e. replying in the same language as the query).

\subsection{Risks and Ethics}
A shortcoming of this work is the lack of focus on second- and third-order interactions, especially those looking at the differences in IFL between languages. The reason choosing to aggregate results in a setting where we expect heterogeneity is the focus of the paper: to define and demonstrate IFL as a problem that is not idiosyncratic to single languages. In future work we hope to better understand the heterogeneous patterns of IFL and how it relates both to structural differences between languages and their relative representation in LLM training corpora.

We do not foresee harm stemming from use of the models and data generated for this research. In particular, the models were trained for the purpose of assessing training decisions, and are unlikely to be competitive with other publicly available models engineered for optimal performance. Nevertheless, the model releases will be accompanied with clear caveats for safety.

While this paper has stressed the importance of research on multilingual VLMs for creating globally inclusive technology, we also want to stress the ethical risks of a mode of research that externalizes the contextualization and nuance necessary to achieve truly inclusive goals. Although the authors of this paper come from a broad constituency of cultures and languages, several of the language used in this study are not spoken by any of the authors. We take seriously the challenge of contributing to a global academic ideal while not appropriating languages or cultures, and will do our best to address any oversights in this respect.


\section*{Acknowledgements}
Authors MH, MLO, SY, AB, ST and VL are employees of the Intel Corporation.
The work of CH and AL is funded under the Excellence Strategy of the German Federal Government and the States.



\bibliography{01-references,styleguide/anthology}

\clearpage
\section*{Appendix}
\appendix
\section{Computational Experiments}

\paragraph{Computational Budget}
The training experiments for this paper were conducted on an internal cluster using nodes with 8 $\times$ A6000 Nvidia 48GB GPUs. In total, we trained 32 distinct configurations (not all of which were ultimately used). A single end-to-end training run with a 7-billion parameter LLM backbone takes 25 hours, meaning roughly 800 GPU hours were used for training. Inference experiments were run on a mixture of RTX3090 24GB cards, A6000 24GB cards and A6000 48GB cards. These required roughly an additional 900 GPU hours. Data analysis utilized CPU. The only sizable compute consisted of applying the DSL estimator to large datasets, which required on the order of $\sim$ 500 CPU hours. Finally, the GPT-4o annotation for the roughly 730k completions in our experiments required roughly USD 2k worth of completion calls.

\section{Expert Annotation}

\paragraph{Sampling Weights}
We stratified on evaluation benchmark (i.e. we weighted the probability by the inverse proportion of the originating benchmark to the full dataset) and then upweighted German by a factor of 4, Chinese and Hindi by 2, and downweighted Romanian, Russian and Urdu by a factor of 2. We sampled a total of $1000$ observations (without replacement) using these weights.

\paragraph{Annotation Procedure}
The $1000$ observations were uploaded into spreadsheets for the authors to manually annotate. Where possible, annotations were matched to authors who could read the language used in the query. The annotation consisted of three questions: what language is the answer, does the model completion match the gold standard, and is the answer correct. The latter two questions were restricted to three categories: true, false and NA. NA was used where the model did not provide coherent output.

\section{Automated Evaluation}

\paragraph{GlotLID} 
We use the GlotLID v3 \citep{kargaran-etal-2023-glotlid} model for automated language identification. We take the most-likely language as predicted by GlotLID, and then manually process the label to collapse what we thought were common misclassifications by the model, such as classifying Mandarin Chinese into various languages and dialects using the simplfied Chinese script when the outputs contained a mix of non-Chinese punctuation characters and Chinese glyphs.

The full parsing rule is as follows:

\begin{verbatim}
def parse_glotlid(lang: str) -> str:
    iso, script = tuple(lang.split("_"))
    match script:
        case "Hani":
            return "chinese"
        case "Jpan":
            return "japanese"
        case "Deva":
            return "hindi"
        case "Beng":
            return "bengali"
        case "Hebr":
            return "hebrew"
        case "Thai":
            return "thai"
        case "Cyrl":
            return "russian"
        case "Zzzz":
            return "none"
        case "Arab":
            match iso:
                case "urd":
                    return "urdu"
                case _:
                    return "arabic"
        case "Latn":
            match iso:
                case "deu":
                    return "german"
                case "eng":
                    return "english"
                case "spa":
                    return "spanish"
                case "ron":
                    return "romanian"
                case "fra":
                    return "french"
                case _:
                    return "other_latin"
        case _:
            return "other"
\end{verbatim}

All designs used the same training parameters as the original LLaVA-v1.5-7B model, and are detailed in the supplementary materials.

Across all pretrained models, we find significant reduction in IFL by interventing on the LLM's intermediate layer. A full breakdown is available in the supplement.

\section{Datasets Used}

Here we provide an overview on the datasets we employ in our study.
\label{sec:related_work_ml_mm_datasets}
\paragraph{MaXM} was introduced by~\citet{changpinyo2023maxm} and is a VQA dataset comprising seven languages in five scripts. In MaXM, the questions and their respective answers are in the same language. Moreover, in MaXM, the images are a subset of the XM3600~\citep{thapliyal2022xm3600} dataset and are chosen to match a region where the language of the question-answer pair is spoken. To increase the cultural diversity, the images selected to match the region where the language of the question-answer pair is spoken.

\paragraph{VisIT-Bench} stands for \textbf{Vis}ual \textbf{I}nstruction \textbf{T}uning \textbf{Bench}mark~\cite{bitton2023visit}. The dataset consists of $592$ vision-language tasks written by human researchers, with GPT-4-generated responses and dense instruction-conditioned captions of the image that are rated by human coders. The $562$ images are taken from the OpenImages~\cite{kuznetsova2020openimages} v7 dataset. In this work we use $525$ examples where the GPT-4 generated responses are rated as correct by human annotators. We machine translate these examples into Arabic, Bengali, Chinese, German, Hebrew, Hindi, Japanese, Spanish and Thai using the Azure Translation API.

%

\paragraph{PALO-LLaVA-Bench-In-The-Wild} dataset is a multilingual VQA dataset created by the PALO authors ~\cite{maaz2024palo} by machine translating the original LLaVA-Bench-In-The-Wild~\cite{liu2023llava} in $10$ languages using a fine-tuned GPT-3.5 instance. The dataset comprises of $60$ questions per language considering $24$ diverse images with a caption describing the visual content. \paragraph{MultiQ} is an evaluation dataset for open-ended question answering covering 137 typologically diverse languages. It is specifically constructed to only contain questions that are simple, factual, and target common knowledge to only test the multilingual capabilities of language models, and no complex reasoning \cite{holtermann2024evaluating}.

\section{Models Used}
Here we provide an overview of the models we employed in our study.

\paragraph{OpenAI/CLIP} is a jointly optimized vision and text feature extractor  trained using large-scale image-caption pairs \cite{radford_learning_2021}. 
CLIP is focused on learning image representations from scratch that are trivially transferable to many downstream tasks without the need for domain specific training. 

\paragraph{DINOv2} is a series of image encoders trained on curated data using unsupervised learning \cite{oquab2024dinov2}. 
Through an improved training recipe and larger dataset, followed by a distillation process of larger to smaller models, DINOv2 is positioned as a ViT-based general-purpose image encoder that surpasses OpenAI/CLIP on most benchmarks.

\paragraph{LLaVA-v1.5} is a large multimodal model trained end-to-end with visual instruction following \cite{liu2023llava}.
The model combines a vision model --- OpenAI/CLIP --- with a large language model --- Vicuna-v1.5 --- achieving impressive visual and language understanding results that were state-of-the-art at its release. 
In this work we used the 7b and 13b variants of the model.

\paragraph{BakLLaVA} is a large multimodal model based on the LLaVA-v1.5 architecture using Mistral-7b as the base LLM \cite{skunkworksai2023bakllava1}.
The model utilizes training data from LLaVA-v1.5 as well as additional sources including ShareGPT and private data with a permissive license.

\paragraph{Yi-6b-chat} is a large language model trained from scratch on English and Chinese corpora \cite{ai2024yi}. 
In this work, we use the 6b variant that has been extended with chat-style training. 

\paragraph{Leo-7b-chat} is a large language model that extends Llama-2 into German through continued training on a large German corpus \cite{pluster2023leolm}. 

\paragraph{GlotLID v3} is a language identification model that coveres 2102 languages. 
The data used to train this model was sourced from Wikipedia, news sites,
translation corpora, religious text, and storybooks. 

\paragraph{NLLB-1.7-distilled} is translation model that support direct translation between 200 languages, including many low-resource languages \cite{nllbteam2022language}. 
The datasets used to train NLLB (No Language Left Behind) were sourced from professionally translated sentences in the Wikipedia domain in addition to publicly available translation datasets. 

\paragraph{GPT-4o} is a commercial large language model provided from OpenAI.

\section{Technical Explainers}

\subsection{Primer on LLaVA}
\paragraph{What is LLaVA?}

Our study analyzes LLaVA, a multimodal model (VLM) that integrates a pretrained vision encoder, denoted as $E_V$, with a large language model (LLM), using a connecting multilayer perceptron (MLP). The process is defined in two main training stages: pretraining of the MLP and joint finetuning of the MLP with the LLM.

\paragraph{Model Architecture}
The VLM comprises the following components:

\textit{Vision Encoder:} The vision encoder $E_V$ processes the visual input $X_v$ to produce a set of embeddings $E_V(X_v)$.

\textit{MLP Connector:} A connecting MLP, defined as $F$, transforms the output of $E_V$ into the dimenstionality of the LLM. This transformation is represented as $F(E_V(X_v))$.

\textit{LLM:} The LLM processes both textual query $X_q$ and the transformed vision embeddings. The combined input to the LLM is given by concatenating the embeddings from the MLP with text embeddings, i.e., $LLM([F(E_V(X_v)); E_L(X_q)])$, where $E_L$ denotes 

The VLM is defined as a function that takes an image input $X_v$ and a textual question $X_q$, and processes these through the vision encoder, MLP connector $F$, and LLM to produce an output $X_a$, which is the model's answer to the question based on the visual context. Formally, the VLM can be expressed as:
\begin{equation}
    VLM(X_v, X_q) = LLM\left([\text{F}(E_V(X_v)); E_L(X_q)]\right),
\end{equation}
where $E_V(X_v)$ is the output of the vision encoder for the input image, $\text{F}(E_V(X_v))$ is the transformed visual embedding suitable for the LLM, and $E_L(X_q)$ represents the embedded form of the textual question. The final output $X_a$ is generated by the LLM, which synthesizes and integrates both the visual and textual information to produce a contextually appropriate answer.

\paragraph{Training Procedure}
The training of the VLM is structured into two distinct stages: pretraining and finetuning.
During the pretraining stage, the MLP is trained while keeping $E_V$ and the LLM frozen. The objective is to optimize the MLP to map the vision encoder outputs to a representation that is effectively integrable with the LLM. The training uses a custom dataset of 595k samples filtered from CC3M \cite{sharma-etal-2018-conceptual}:
\begin{equation}
    \mathcal{L}_{\text{MLP}} = \sum_{(X_v, X_c) \in \mathcal{D}} L_{CE}(VLM(X_v, X_q)),
\end{equation}
where $X_c$ represents the captions associated with $X_v$, and $\mathcal{D}$ denotes the dataset.

\paragraph{Finetuning}
In the finetuning stage, the MLP and the LLM are jointly trained with a larger, diverse dataset of 665k multimodal instruction tuning examples, integrating both synthetic and established vision-language training sets. The entire conversation $C = (X_q, X_a)$ is fed into the LLM, with autoregressive masking applied to focus training on the answers using supervised cross-entropy loss $L_{CE}$:
\begin{equation}
    \mathcal{L}_{\text{VLM}} = \sum_{C \in \mathcal{C}} L_{CE}(VLM(X_v, X_q)),
\end{equation}
where $\mathcal{C}$ represents the conversation dataset, and training focuses exclusively on the answer parts $X_a$, leveraging the context provided by the entire conversation but training only through the answer segments.

\begin{table}
\begin{tabular}{lllll}
llava7b    &          &        &              &            \\
\toprule
dataset    &   Lang.  & IFL    & \begin{tabular}[c]{@{}c@{}}IFL + \\ Remedy\end{tabular} & Diff. \\ \midrule
llavaw     & ar     & -0.250 & -0.083       & 0.167      \\
           & bn     & -0.117 & -0.050       & 0.067      \\
           & zh     & -0.233 & -0.017       & 0.217      \\
           & fr     & -0.183 & 0.000        & 0.183      \\
           & hi     & -0.133 & -0.033       & 0.100      \\
           & ja     & -0.117 & -0.050       & 0.067      \\
           & ru     & -0.233 & -0.017       & 0.217      \\
           & es     & -0.200 & -0.050       & 0.150      \\
           & ur     & -0.050 & 0.083        & 0.133      \\ \hline
maxm       & zh     & 0.004  & 0.000        & -0.004     \\
           & fr     & 0.004  & -0.011       & -0.015     \\
           & he     & -0.132 & -0.125       & 0.007      \\
           & hi     & -0.042 & -0.035       & 0.008      \\
           & ro     & 0.000  & -0.014       & -0.014     \\
           & th     & -0.007 & -0.011       & -0.004     \\ \hline
visitazure & ar     & -0.038 & -0.047       & -0.009     \\
           & bn     & -0.084 & -0.038       & 0.045      \\
           & zh     & -0.026 & -0.047       & -0.021     \\
           & de     & -0.054 & -0.037       & 0.017      \\
           & he     & -0.038 & -0.037       & 0.002      \\
           & hi     & -0.026 & -0.009       & 0.017      \\
           & ja     & -0.021 & -0.051       & -0.030     \\
           & es     & -0.024 & -0.042       & -0.017     \\
           & th     & -0.045 & -0.010       & 0.035      \\ \hline
average    &  -     & -0.085 & -0.030       & 0.055     \\
\bottomrule
\end{tabular}
\caption{mechint raw llava7b scores. }
    \label{tab:mechint_raw_llava7b}
\end{table}

\begin{table}
\begin{tabular}{ccccc}
llava13b   &          &        &              &            \\
\toprule
dataset    &   Lang.  & IFL    & \begin{tabular}[c]{@{}c@{}}IFL + \\ Remedy\end{tabular} & Diff. \\ \midrule
llavaw     & ar     & -0.183 & -0.033       & 0.150      \\
           & bn     & -0.233 & -0.083       & 0.150      \\
           & zh     & -0.133 & -0.033       & 0.100      \\
           & fr     & -0.200 & -0.100       & 0.100      \\
           & hi     & -0.317 & -0.200       & 0.117      \\
           & ja     & -0.183 & -0.117       & 0.067      \\
           & ru     & -0.433 & -0.317       & 0.117      \\
           & es     & -0.233 & -0.183       & 0.050      \\
           & ur     & -0.550 & -0.267       & 0.283      \\ \hline
maxm       & zh     & -0.025 & -0.007       & 0.018      \\
           & fr     & -0.008 & -0.045       & -0.038     \\
           & he     & -0.175 & -0.121       & 0.054      \\
           & hi     & -0.042 & -0.035       & 0.008      \\
           & ro     & -0.106 & -0.085       & 0.021      \\
           & th     & -0.157 & -0.093       & 0.063      \\ \hline
visitazure & ar     & -0.174 & -0.066       & 0.108      \\
           & bn     & -0.244 & -0.136       & 0.108      \\
           & zh     & -0.071 & -0.031       & 0.040      \\
           & de     & -0.105 & -0.094       & 0.010      \\
           & he     & -0.125 & -0.082       & 0.044      \\
           & hi     & -0.136 & -0.096       & 0.040      \\
           & ja     & -0.056 & -0.044       & 0.012      \\
           & es     & -0.057 & -0.042       & 0.016      \\
           & th     & -0.258 & -0.155       & 0.103      \\ \hline
average    &  -     & -0.175 & -0.103       & 0.073      \\ 
\bottomrule
\end{tabular}
\caption{Mechanistic intervention complete llava13b scores. }
    \label{tab:mechint_raw_llava13b}
\end{table}

\begin{table}
\begin{tabular}{ccccc}
bakllava   &          &        &              &            \\
\toprule
dataset    &   Lang.  & IFL    & \begin{tabular}[c]{@{}c@{}}IFL + \\ Remedy\end{tabular} & Diff. \\ \midrule
llavaw     & ar     & 0.000  & 0.350        & 0.350      \\
           & bn     & -0.050 & 0.217        & 0.267      \\
           & zh     & -0.033 & -0.067       & -0.033     \\
           & fr     & -0.117 & 0.000        & 0.117      \\
           & hi     & 0.000  & 0.050        & 0.050      \\
           & ja     & -0.017 & -0.067       & -0.050     \\
           & ru     & 0.000  & 0.000        & 0.000      \\
           & es     & -0.117 & 0.217        & 0.333      \\
           & ur     & -0.017 & 0.183        & 0.200      \\ \hline
maxm       & zh     & -0.018 & 0.014        & 0.032      \\
           & fr     & -0.318 & -0.223       & 0.095      \\
           & he     & 0.000  & 0.029        & 0.029      \\
           & hi     & 0.000  & 0.135        & 0.135      \\
           & ro     & -0.567 & -0.299       & 0.268      \\
           & th     & -0.119 & -0.078       & 0.041      \\ \hline
visitazure & ar     & -0.010 & 0.608        & 0.618      \\
           & bn     & -0.012 & 0.557        & 0.570      \\
           & zh     & -0.007 & 0.019        & 0.026      \\
           & de     & -0.136 & -0.108       & 0.028      \\
           & he     & 0.000  & 0.078        & 0.078      \\
           & hi     & -0.007 & 0.113        & 0.120      \\
           & ja     & -0.014 & 0.026        & 0.040      \\
           & es     & -0.183 & 0.291        & 0.474      \\
           & th     & -0.007 & 0.294        & 0.301      \\ \hline
average    &  -     & -0.073 & 0.098        & 0.170      \\ 
\bottomrule
\end{tabular}
\caption{Mechanistic intervention complete bakllava scores. }
    \label{tab:mechint_raw_bakllava}
\end{table}

\begin{table}
\begin{tabular}{ccccc}
\multicolumn{2}{c}{llavagemma2b} &        &              &            \\
\toprule
dataset    &   Lang.  & IFL    & \begin{tabular}[c]{@{}c@{}}IFL + \\ Remedy\end{tabular} & Diff. \\ \midrule
llavaw           & ar     & -0.583 & -0.533       & 0.050      \\
                 & bn     & -0.483 & -0.483       & 0.000      \\
                 & zh     & -0.733 & -0.567       & 0.167      \\
                 & fr     & -0.800 & -0.433       & 0.367      \\
                 & hi     & -0.500 & -0.317       & 0.183      \\
                 & ja     & -0.600 & -0.650       & -0.050     \\
                 & ru     & -0.883 & -0.650       & 0.233      \\
                 & es     & -0.900 & -0.700       & 0.200      \\
                 & ur     & -0.517 & -0.483       & 0.033      \\ \hline
maxm             & zh     & -0.852 & -0.762       & 0.090      \\
                 & fr     & -0.905 & -0.652       & 0.254      \\
                 & he     & -0.768 & -0.482       & 0.286      \\
                 & hi     & -0.731 & -0.546       & 0.185      \\
                 & ro     & -0.810 & -0.637       & 0.173      \\
                 & th     & -0.646 & -0.455       & 0.190      \\ \hline
visitazure       & ar     & -0.718 & -0.578       & 0.139      \\
                 & bn     & -0.483 & -0.420       & 0.063      \\
                 & zh     & -0.672 & -0.552       & 0.120      \\
                 & de     & -0.688 & -0.258       & 0.430      \\
                 & he     & -0.617 & -0.280       & 0.336      \\
                 & hi     & -0.589 & -0.375       & 0.214      \\
                 & ja     & -0.526 & -0.509       & 0.017      \\
                 & es     & -0.793 & -0.608       & 0.185      \\
                 & th     & -0.538 & -0.373       & 0.166      \\ \hline
 average         &  -     & -0.681 & -0.513       & 0.168      \\ \toprule
\end{tabular}
\caption{Mechanistic intervention complete LLaVA-Gemma-2b scores. }
    \label{tab:mechint_raw_llava_gemma}
\end{table}


\begin{figure*}[ht]
\centering
\includegraphics[width=1.\textwidth]{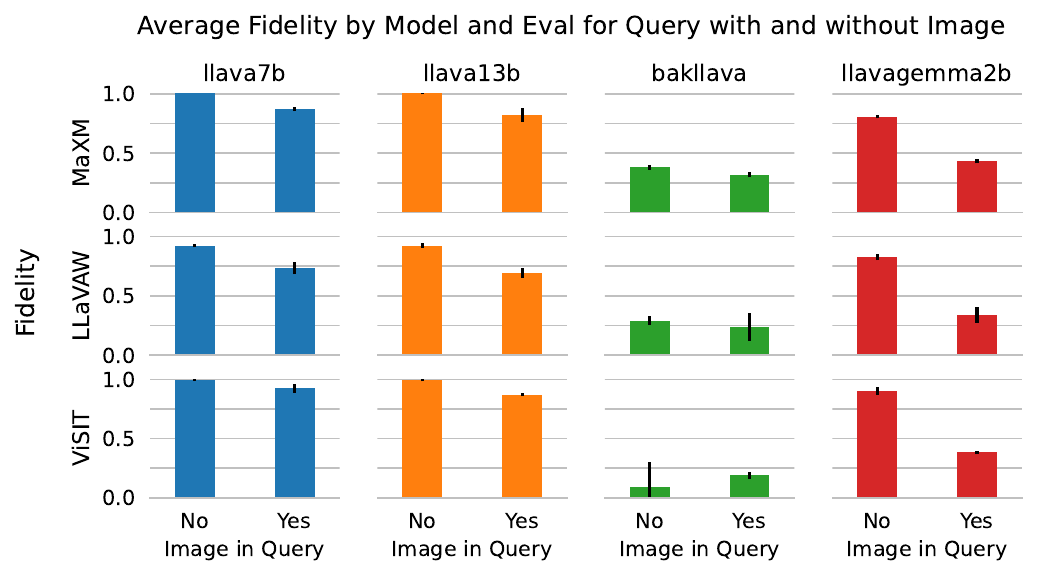}
\caption{ Average fidelity by model and eval for query with and without Images}
\end{figure*}

\section{Use of AI Tools}
The authors of this paper used Github Co-pilot for coding assistance for this research.






\end{document}